\documentclass[journal]{IEEEtran}
\pdfoutput=1
\usepackage{lineno,hyperref}
\modulolinenumbers[5]

\usepackage{multirow}
\usepackage{amsmath}
\usepackage{algorithm} 
\usepackage{algpseudocode}
\usepackage{amsmath}
\usepackage{algorithm}
\usepackage{graphicx}

\usepackage{xspace}

\makeatletter
\DeclareRobustCommand\onedot{\futurelet\@let@token\@onedot}
\def\@onedot{\ifx\@let@token.\else.\null\fi\xspace}

\makeatother

\newcommand{\colorcomment}[3]{\xspace{\color{#2}[{#1}]:{#3}}\xspace}

\newcommand{\byung}[1]{\colorcomment{Byung}{magenta}{#1}}

\newcommand{\cmmnt}[1]{\ignorespaces}

\algnewcommand\algorithmicforeach{\textbf{for each}}
\algdef{S}[FOR]{ForEach}[1]{\algorithmicforeach\ #1\ \algorithmicdo}

\usepackage{color}
\usepackage{graphicx}
\usepackage{verbatim}

\def\BibTeX{{\rm B\kern-.05em{\sc i\kern-.025em b}\kern-.08em
  T\kern-.1667em\lower.7ex\hbox{E}\kern-.125emX}}
\begin{document}

\title{Graph Anomaly Detection with Graph Neural Networks: Current Status and Challenges}

\author{\IEEEauthorblockN{Hwan Kim,
Byung Suk Lee,
Won-Yong Shin,~\IEEEmembership{Senior Member,~IEEE}, and
Sungsu Lim,~\IEEEmembership{Member,~IEEE}}
\IEEEcompsocitemizethanks{\IEEEcompsocthanksitem H. Kim and S. Lim are with the Department of Computer Science and Engineering, Chungnam National University, Daejeon 34134, Republic of Korea (e-mail: hwan.kim@o.cnu.ac.kr, sungsu@cnu.ac.kr).
\IEEEcompsocthanksitem B. S. Lee is with the Department of Computer Science, University of Vermont, Burlington, Vermont 05405, USA (e-mail: bslee@uvm.edu).
\IEEEcompsocthanksitem W.-Y. Shin is with the School of Mathematics and Computing (Computational Science and Engineering), Yonsei University, Seoul 03722, Republic of Korea, and also with Pohang University of Science and Technology (POSTECH) (Artificial Intelligence), Pohang 37673, Republic of Korea (e-mail: wy.shin@yonsei.ac.kr).}
\thanks{{\it (Corresponding author: Sungsu Lim.)}}}

\maketitle

\begin{abstract}
Graphs are used widely to model complex systems, and detecting anomalies in a graph is an important task in the analysis of complex systems. Graph anomalies are patterns in a graph that do not conform to normal patterns expected of the attributes and/or structures of the graph. In recent years, graph neural networks (GNNs) have been studied extensively and have successfully performed difficult machine learning tasks in node classification, link prediction, and graph classification thanks to the highly expressive capability via message passing in effectively learning graph representations. To solve the graph anomaly detection problem, GNN-based methods leverage information about the graph attributes (or features) and/or structures to learn to score anomalies appropriately. In this survey, we review the recent advances made in detecting graph anomalies using GNN models. Specifically, we summarize GNN-based methods according to the graph type (i.e., static and dynamic), the anomaly type (i.e., node, edge, subgraph, and whole graph), and the network architecture (e.g., graph autoencoder, graph convolutional network). To the best of our knowledge, this survey is the first comprehensive review of graph anomaly detection methods based on GNNs.
\end{abstract}

\begin{IEEEkeywords}
Dynamic graph, graph anomaly detection, graph neural network, node anomaly, static graph.
\end{IEEEkeywords}

\section{Introduction}
\IEEEPARstart{G}{raphs} are an effective data structure for efficiently representing and extracting complex patterns of data and are used widely in numerous areas like social media, e-commerce, biology, academia, communication, and so forth.
Data objects represented in a graph are interrelated, and the objects are typically represented as nodes and their relationships as edges between nodes. 
The structure of a graph refers to how the nodes are related via individual edges, and  
can effectively represent even far-reaching relationships between nodes.
Moreover, graphs can be enriched semantically by augmenting their structural representations with attributes of nodes and/or edges.
Anomaly detection is the process to identify abnormal patterns that significantly deviate from patterns that are typically observed. 
This is an important task with increasing needs and applications in various domains.
There have been significant research efforts on anomaly detection since Grubbs et al. \cite{grubbs1969procedures} first introduced the notion of anomaly (or outlier). 
Since then, with the advancement of graph mining over the past years,
graph anomaly detection has been drawing much attention \cite{ranshous2015anomaly,akoglu2015graph}.

Early work on graph anomaly detection has been largely dependent on domain knowledge and statistical methods, where features for detecting anomalies
have been mostly handcrafted.
This handcrafted detection task is naturally very time-consuming and labor-intensive. 
Furthermore, real-world graphs often contain a very large number of nodes and edges labeled with a large number of attributes, and are thus large-scale and high-dimensional. 
 %
 %
To overcome the limitations of the early work, considerable attention has been paid to 
deep learning approaches recently when detecting anomalies from graphs \cite{li2019deep}. 
Deep learning's 
multi-layer structure with non-linearity can examine large-scale high-dimensional data and extract patterns from the data, thereby achieving satisfactory performance without the burden of handcrafting features \cite{wang2019deep,langkvist2014review}.

\begin{figure}[t!]
    \centering
    \includegraphics[width=\columnwidth]{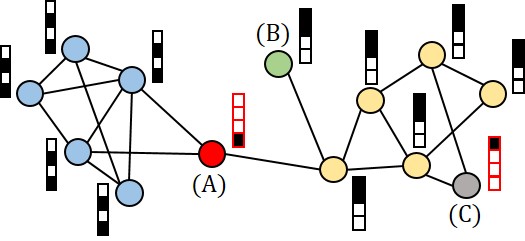}
    \caption{Example of graph anomaly detection. Nodes A and C are detected anomalous attribute-wise. Nodes A and B are detected anomalous structure-wise (as they do not belong to any community). Using GNN detects node A to be anomalous both attribute-wise and structure-wise.}
    \label{fig:Anomaly_type}
\end{figure}

More recently, graph neural networks (GNNs) have been adopted to efficiently and intuitively detect anomalies from graphs due to the highly expressive capability via the message passing mechanism in learning graph representations (e.g., \cite{ding2019deep,li2019specae}).
With GNNs, learning and extracting anomalous patterns from graphs, even those with highly complex structures or attributes, are relatively straightforward as GNN itself handles a graph with attributes as the input data~\cite{wu2020comprehensive}. 
The state-of-the-art graph anomaly detection approaches \cite{ding2019deep,zheng2019addgraph}
combine GNN with existing deep learning approaches, in which GNN captures the characteristics of a graph and deep learning captures other types of information (e.g., time).
  %
Fig. \ref{fig:Anomaly_type} illustrates an example of graph anomaly detection with GNN. Suppose that nodes (A) and (C) are detected anomalous in terms of the node attributes, and nodes (A) and (B) are detected anomalous in terms of the graph topology. Then, only node (A) should be detected anomalous if both node attributes and graph topology are taken into account together as anomalous factors. GNN models enable us to detect such anomalies by examining both graph topology and node attributes simultaneously.

In this survey, we provide an overview of GNN-based approaches for graph anomaly detection and review them primarily by the types of graphs, namely static graphs and dynamic graphs.
Compared with other surveys on related topics --- on graph anomaly detection (in general) \cite{ranshous2015anomaly,akoglu2015graph}, graph anomaly detection specifically using deep learning \cite{kwon2019survey,ma2021comprehensive}, and general anomaly detection using deep learning \cite{pang2021deep,chalapathy2019deep} --- 
 %
this survey aims to touch on the unique angle of graph anomaly detection \emph{using GNN models}. 
Given the significance of GNN and the active ongoing research efforts for its use in graph anomaly detection, it is our conviction that a comprehensive survey on this particular topic is timely and beneficial to the research community.

\begin{figure*}[ht]
    \centering
    \includegraphics[width=\textwidth]{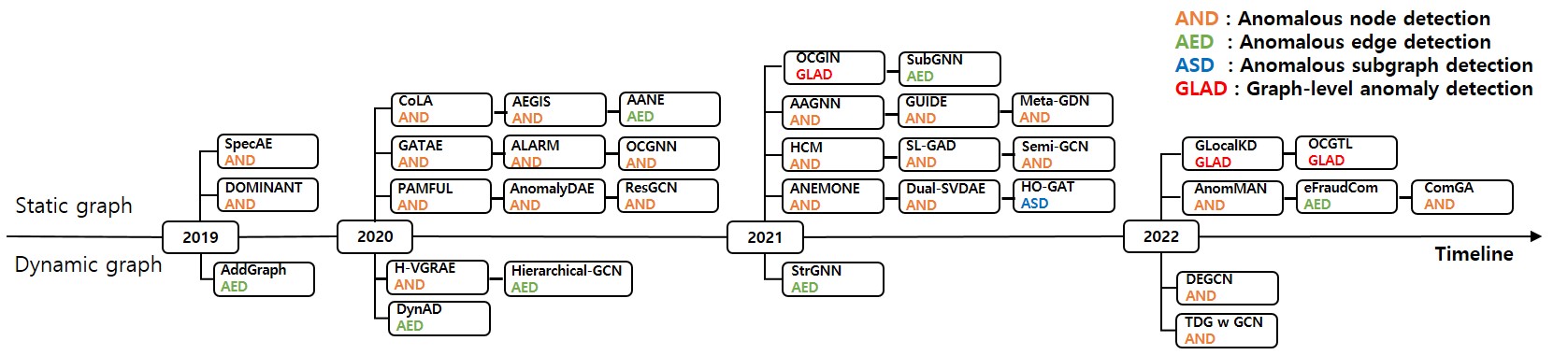}
    \caption{Timeline of graph anomaly detection methods using GNN models. The methods above the timeline are for static graphs, and those below are for dynamic graphs.}
    \label{fig:Timeline}
\end{figure*}

Fig. \ref{fig:Timeline} shows the timeline of the surveyed methods. The survey in Section \ref{sec:GNNGAD} is organized according to the classification framework used by other surveys on graph anomaly detection \cmmnt{(in general, not necessarily using GNN)} \cite{akoglu2015graph,ranshous2015anomaly,kwon2019survey,ma2021comprehensive}. This area is still new, and yet the published methods cover a broad range of graphs (static versus dynamic, plain versus attributed) and anomaly types (structure, node, edge, subgraph) although the distribution of the study topics seems skewed toward node anomalies in static graphs. Additionally, in Section \ref{sec:oppchal}, we share our opinions on several promising opportunities and challenges pertaining to graph anomaly detection using GNN. 

 %



\section{GNN Frameworks}

GNNs have been widely used as an effective method to extract useful features, especially from attributed networks, while performing graph representation learning (also known as network embedding) through the message passing mechanism.
 %
Model architectures of GNNs have been actively developed, and include, but not limited to, graph convolutional network (GCN) \cite{kipf2016semi}, GraphSAGE \cite{hamilton2017inductive}, and graph attention network (GAT) \cite{velivckovic2017graph}.
 %
The main idea of such GNN models is message passing, which aggregates individual features from $h$-hop neighbors.
The message passing mechanism of GNNs can be expressed as follows~\cite{hamilton2020graph}:
\begin{align*}
   \mathbf{m}_{u}^{(k-1)}&\leftarrow\textrm{AGGREGATE}^{(k-1)} (\{ \mathbf{h}_v^{(k-1)}, \forall v \in \mathcal{N}(u) \cup \{u\} \}),\\
   \mathbf{h}_u^{(k)}&\leftarrow\textrm{UPDATE}^{(k-1)}\left( \mathbf{h}_u^{(k-1)}, \mathbf{m}_{u}^{(k-1)} \right),
\end{align*}
where $\textrm{AGGREGATE}$ and $\textrm{UPDATE}$ are arbitrary differentiable functions and $\mathbf{m}_{u}$ is the message that is aggregated from the neighborhood $\mathcal{N}(u)$ of node $u$ as well as the node itself.
At the $k$-th layer of a GNN, the previous embedding $\mathbf{h}_u^{(k-1)}$ of $u$ is updated to $\mathbf{h}_u^{(k)}$ by aggregating neighborhood information via message passing.
The initial embeddings are set to the features of nodes, and the output at the final layer corresponds to the embeddings of the underlying GNN model.

GCN originally presented an effective network representation model that naturally combines the network structure and node attributes in the
learning process.
 %
Subsequently, Hamilton et al.~\cite{hamilton2017inductive} introduced GraphSAGE, which samples nodes for message aggregation in a neighbor to handle large graphs.
GAT learns the weights of each node in a neighbor during message passing.


There have been follow-up studies that use alternative model architectures. 
 %
 %
For example, Wang et al.~\cite{wang2020graph} introduced a graph stochastic neural network (GSNN) to alleviate the inefficiency of a deterministic way used in existing GNN models.
Most existing GNN approaches embed networks into either the Euclidean or hyperbolic space. These approaches have the issue of inflexibility in modeling a complex graph topology. To address this issue, Zhu et al.~\cite{zhu2020graph} proposed a graph geometry interaction learning (GIL) algorithm, which employs \emph{both} Euclidean and hyperbolic spaces.

\section{GNN-Based Graph Anomaly Detection}\label{sec:GNNGAD}
The surveyed methods are categorized by the graph types (static versus dynamic), anomaly types (nodes, edges, subgraphs, and whole graphs), and network architectures. Table \ref{tab:method summary} outlines the organization of our survey.

\subsection{GNN-based static graph anomaly detection}

A majority of research efforts on static graph anomaly detection addressed node anomalies, and only a few addressed edge anomalies and subgraph anomalies.  
%

\subsubsection{Anomalous Node Detection}\label{sec:ANDS}

Detecting anomalous \emph{nodes} using GNNs was carried out mostly in attributed graphs. That is, each of GNN-based methods extracts node attribute information as well as structural information from a static attributed graph and evaluates the anomaly score of nodes using a certain scoring algorithm. Various GNN-based approaches have been proposed to effectively extract the necessary features from attributed graphs. We categorize these methods according to their network architectures and then briefly describe the key ideas behind each method. 

On the basis of the structural information, anomalous nodes can be further divided into the following three types: global anomalies, structural anomalies, and community anomalies \cite{ma2021comprehensive}.
{\em Global} anomalies are referred to as deviated node attributes in the graph; 
{\em structural} anomalies are referred to as deviated structural information in the graph; 
and {\em community} anomalies are referred to as both deviated node attributes and structural information in the same community. 

Note that a fair number of GNN-based approaches are built upon the graph autoencoder (GAE) framework instantiated with either GCN or GAT. For anomalous node detection in static graphs, we review GNN-based GAE methods as well as standalone GNN methods.

\begin{table*}
\centering
\caption{Taxonomy and summary of graph anomaly detection methods using GNN models. 
}
\label{tab:method summary}
\hspace*{-1em}
\resizebox{1.025\textwidth}{!}{%
\begin{tabular}{|c|c|c|c|c|}
\hline
Graph type & Anomaly type & Network architecture                   & Method           & Summary (key issue addressed $\to$ solution)                                                   \\ \hline \hline
\multirow{27}{*}{Static graph} &  \multirow{20}{*}{Node anomaly} &  \multirow{8}{*}{GCN-based GAE} &  DOMINANT~\cite{ding2019deep} (2019) &  Complex interactions, sparsity, non-linearity $\to$ GCN-based encoder \\ \cline{4-5} 
           &              &                                        & Dual-SVDAE~\cite{zhang2021deep} (2021) & Overfitting for normal \& abnormal $\to$ hypersphere embedding space                 \\ \cline{4-5} 
           &              &                                        & GUIDE~\cite{yuan2021higher} (2021)    & Complex interactions $\to$ higher-order structure decoder                                          \\ \cline{4-5} 
           &              &                                        & SpecAE~\cite{li2019specae} (2019)  & Over-smoothing issue $\to$ tailored embedding space                                         \\ \cline{4-5} 
           &              &                                        & ComGA~\cite{luo2022comga} (2022)    & Over-smoothing issue $\to$ community-specific representation             \\ \cline{4-5} 
           &              &                                        & ALARM~\cite{peng2020deep} (2020)   & Heterogeneous attributes $\to$ multiple GCN-based encoders                            \\ \cline{4-5} 
           &              &                                        & AnomMAN~\cite{chen2022anomman} (2022) & Heterogeneous attributes $\to$ multiple GCN-based encoders                           \\ \cline{4-5} 
           &              &                                        & SL-GAD~\cite{zheng2021generative} (2021)  & Contextual information $\to$ subgraph sampling \& contrastive learning         \\ \cline{3-5} 
           &              & \multirow{6}{*}{GCN alone}             & Semi-GCN~\cite{kumagai2021semi} (2021) & Label information $\to$ semi-supervised learning by GCN                                        \\ \cline{4-5} 
           &              &                                        & HCM~\cite{huang2021hop} (2021)      & Label \& contextual information $\to$ hop-count prediction model                          \\ \cline{4-5} 
           &              &                                        & ResGCN~\cite{pei2021resgcn} (2021)           & Over-smoothing issue $\to$ GCN with residual-based attention                                          \\ \cline{4-5} 
           &              &                                        & CoLA~\cite{liu2021anomaly} (2021)            & Targeting issue of GAE $\to$ contrastive self-supervised learning        \\ \cline{4-5} 
           &              &                                        & ANEMONE~\cite{jin2021anemone} (2021)         & Contextual information $\to$ multi-scale contrastive learning                                             \\ \cline{4-5} 
           &              &                                        & PAMFUL~\cite{zhao2021synergistic} (2021)          & Contextual information $\to$ pattern mining algorithm with GCN                            \\ \cline{3-5} 
           &              & \multirow{3}{*}{GAT-based GAE}         & AnomalyDAE~\cite{fan2020anomalydae} (2020)       & Complex interactions $\to$ GAT-based encoder                                           \\ \cline{4-5} 
           &              &                                        & GATAE~\cite{you2020gatae} (2020) & Over-smoothing issue $\to$ GAT-based encoder                                         \\ \cline{4-5} 
           &              &                                        & AEGIS~\cite{ding2020inductive} (2020)           & Handling unseen nodes $\to$ generative adversarial learning with GAE                        \\ \cline{3-5} 
           &              & \multirow{3}{*}{Other GNN-based model} & OCGNN~\cite{wang2021one} (2021)           & Targeting issue of GAE $\to$ GNN with hypersphere embedding space                                        \\ \cline{4-5} 
           &              &                                        & AAGNN~\cite{zhou2021subtractive} (2021)           & Targeting issue of GAE $\to$ GNN with hypersphere embedding space \\ \cline{4-5} 
           &              &                                        & Meta-GDN~\cite{ding2021few} (2021)         & Hard work to label anomalies $\to$ meta-learning with auxiliary graphs                                     \\ \cline{2-5} 
	   &  \multirow{3}{*}{Edge anomaly} &  \multirow{2}{*}{GCN-based GAE} &  AANE~\cite{duan2020aane} (2020) &  Noise or adversarial links $\to$ GAE with a loss for anomalous links \\ \cline{4-5}
           &                                &                                        & eFraudCom~\cite{zhang2022efraudcom} (2022)        & {Fraud detection $\to$ heterogeneous graph and representative data sampling}                                \\ \cline{3-5} 
           &                                & GCN alone                              & SubGNN~\cite{song2021subgraph} (2021)  & {Fraud detection $\to$ GIN and extracting and relabeling subgraphs} \\ \cline{2-5}
	   & Subgraph anomaly  & GAT-based GAE    & HO-GAT~\cite{huang2021hybrid} (2021)           & Abnormal subgraphs $\to$ hybrid-order attention with motif instances          \\ \cline{2-5}
	                                  & \multirow{3}{*}{\begin{tabular}[c]{@{}c@{}}Graph-level \\ anomaly\end{tabular}} & \multirow{3}{*}{GCN alone}             & OCGIN~\cite{zhao2021using} (2021) & {Graph-level anomaly detection $\to$ graph classification with GIN} \\ \cline{4-5} 
                               &                                                                                 &                  & OCGTL~\cite{qiu2022raising} (2022) & {Hypersphere collapse $\to$ set of GNNs for embedding} \\ \cline{4-5}                                                               
		&	&   & GLocalKD~\cite{ma2021deep} (2022) &  Graph-level anomalies $\to$ joint learning global \& local normality 
		\\ \hline
\multirow{7}{*}{Dynamic graph} &
  \multirow{4}{*}{Edge anomaly} &
  \multirow{4}{*}{GCN and GRU} &  AddGraph~\cite{zheng2019addgraph} (2019) &  Long-term patterns $\to$ temporal GCN with attention-based GRU \\ \cline{4-5} 
           &              &                                        & DynAD~\cite{zhu2020flexible} (2020)           & Long-term patterns $\to$ temporal GCN with attention-based GRU     \\ \cline{4-5} 
           &              &                                        & Hierarchical-GCN~\cite{wang2020hierarchical} (2020) & Dynamic data evaluation $\to$ temporal \& hierarchical GCN               \\ \cline{4-5} 
           &              &                                        & StrGNN~\cite{cai2021structural} (2021)           & Structural change $\to$ mining unusual temporal subgraph structures                                   \\ \cline{2-5} 
 &
  \multirow{3}{*}{Node anomaly} &  GCN \& DRNN-based GAE &  H-VGRAE~\cite{yang2020h} (2020) &  
  Anomalous nodes $\to$ modeling stochasticity and multi-scale ST dependency
  \\ \cline{3-5}
                               &                                                                                 & GCN and GRU                            & DEGCN~\cite{zhang2022malware} (2022)           & {To capture node- and global-level patterns $\to$ DGCN and GGRU}                                                                \\ \cline{3-5} 
                               &                                                                                 & GCN alone                              & TDG with GCN~\cite{zola2022network} (2022)    & {Malicious connections on traffic $\to$ extracting TDGs} \\ \hline

\end{tabular}%
}
\end{table*}


\paragraph{GCN-Based GAE Framework} 

GAE has been most widely used for detecting graph anomalies. 
For anomalous node detection in static graphs, while existing GAEs mostly use GCN in the encoder,
they adopt their own decoders depending on 
the perspective on which each method focuses. Additionally, an anomaly scoring function, following the decoder, identifies the abnormality by scoring each node on the basis of reconstruction errors from the decoder.


It is likely that graphs are sparse in real-world situations.
Moreover, complex interactions between individual nodes are difficult to capture due to the non-linear characteristics.
To address these issues, \emph{deep anomaly detection on attributed networks (DOMINANT)} \cite{ding2019deep} detected 
the global and structural anomalies using GCN as the encoder while discovering node representations from a given attributed graph.
Decoders in DOMINANT were designed in the sense of reconstructing the original graph structure and nodal attributes.


To capture complex interactions of GAE-based approaches in high-dimensional graphs, Zhang et al.~\cite{zhang2021deep} proposed a one-class classification-based framework, called \emph{dual support vector data description based autoencoder (Dual-SVDAE)}, which aims to learn the hypersphere boundary on the normal nodes' latent space based on the graph structure and attributes in order to detect global and structural anomalies. Specifically, its AE has a dual encoder--decoder architecture --- 1) a GCN-based encoder and an inner product-based decoder for structural features and 2) a multilayer perceptron (MLP)-based encoder--decoder for attributes.
The nodes outside of the hypersphere space are regarded as anomalous nodes.


Subsequently, Yuan et al.~\cite{yuan2021higher} presented a dual GAE framework, \emph{higher-order structure based anomaly detection (GUIDE)}, which leverages the higher-order structures in modeling the complex interaction to detect global and structural anomalies.
Specifically, it consists of the attribute AE for attribute embeddings and the structure AE for structural embeddings based on the attention mechanism.


Since GCNs learn node representations via message passing from the neighbors, they could over-smooth the representations, making anomalous nodes less distinguishable from the normal nodes.
To alleviate the over-smoothing issue, \emph{spectral AE (SpecAE)} \cite{li2019specae} was presented an approach for detecting global and community anomalies by employing the Laplacian to calculate the inter-node distances and to simultaneously embed attributes and relations in the underlying graph. It used the GCN encoder and the deconvolutional decoder.
Similarly, Luo et al.~\cite{luo2022comga} proposed \emph{community-aware attributed graph anomaly detection framework (ComGA)} to detect global and community anomalies by propagating community-specific representations of each node with a gateway.
In a community detection module, communities were encoded and decoded for community representations. In a tailored deep GCN (tGCN) module, the representations, structural information, and nodal attributes were used as input of GCN layers. The gateway concatenates the community feature, structure, and attribute. Finally, in an anomaly detection module, the well-designed structure decoder and attribute decoder reconstruct the output from the tGCN module.


\cmmnt{On the other hand, }Most existing studies do not carefully take multi-view properties into consideration.
The complementary information from multi-view attributes can help detect anomalies more efficiently. In a multi-view attributed network, 
each node is affiliated with a set of multi-dimensional features (attributes), which can be represented by $k$ distinct feature spaces along with $k$ views.
By making use of the multi-view attributes,
Peng et al.~\cite{peng2020deep} proposed a \emph{deep multi-view framework for anomaly detection (ALARM)} for detecting global and structural anomalies.
Specifically, it employs multiple GCNs to embed a multi-view attributed graph as an encoder and uses two decoders, each of which reconstructs the graph structure and attributes.

Chen et al.~\cite{chen2022anomman} proposed a \cmmnt{GCN-based GAE} framework, called \emph{anomaly on multi-view attributed networks on attributed networks (AnomMAN)}, which aims to detect community anomalies.
It first decomposes multi-view attributed graphs into $k$-subgraphs and encodes the subgraphs with GCNs. Then, the representations from different subgraphs are concatenated using the attention mechanism and reconstructed by a decoder with respect to both structures and attributes of the multi-view graphs.


The existing GAE-based approaches do not effectively extract the contextual information (e.g., neighboring nodes and subgraphs).
In addition, they focused primarily on node-level representation learning.
To address these issues, Zheng et al.~\cite{zheng2021generative} proposed \emph{self-supervised learning for graph anomaly detection (SL-GAD)}, where the GCN-based encoder and decoder are used to capture node- and subgraph-level features along with a self-supervised learning strategy for community anomalies.
Specifically, a target node and subgraphs centered on each target node (i.e., contextual information) were sampled and fed into the encoder.
Afterward, generative attribute reconstruction and contrastive learning modules were leveraged to extract the useful features in a self-supervised fashion.
The GCN-based decoder reconstructs the node attributes, and then the anomaly score is calculated on the basis of both modules.

\paragraph{GCN Framework}


Since labeling anomalies is labor-intensive and costly, many efforts have focused on unsupervised learning.
Nonetheless, labeling information can help enhance the model performance.
For this reason, Kumagai et al.~\cite{kumagai2021semi} proposed a simple yet effective GCN-based method, called \emph{semi-GCN}, which is capable of embedding nodes into a hypersphere space while taking advantage of the structure and attribute features of a graph by stacking graph convolutional layers in order to detect global anomalies.
%


To address the issue above in unsupervised learning and leverage global context information, 
Huang et al.~\cite{huang2021hop} proposed the \emph{hop-count based model (HCM)} based on self-supervised learning with the \emph{PairwiseDistance} algorithm~\cite{jin2020self} in which hop counts of node pairs are leveraged as labels \cite{jin2020self}.
Specifically, it consists of a preprocessing module for labeling, GCNs for learning the representation of graphs, and the MLP module for calculating the anomaly scores.


Pei et al.~\cite{pei2021resgcn} proposed \emph{residual GCN (ResGCN)} to alleviate the sparsity and over-smoothing issues in modeling attributed graphs. It uses the attention mechanism with residual information, modeled by MLP layers, in a graph.


Basically, GAE can bring sub-optimal performance on the anomaly detection task.
To tackle this problem raised by GAE-based methods, Liu et al.~\cite{liu2021anomaly} proposed \emph{contrastive self-supervised learning (CoLA)}, which concentrates on modeling the relations between nodes and their neighboring subgraphs to detect the community anomalies.
More precisely, the well-organized node and its subgraph pairs were carefully sampled, and then the GCN-based contrastive learning model embedded the pairs.
Subsequently, a readout module computed discriminative scores, while showing the abnormality of a target node of the pairs.


Jin et al.~\cite{jin2021anemone} presented \emph{graph anomaly detection framework with multi-scale contrastive learning (ANEMONE)}, which detects community anomalies, to overcome the sub-optimality based on a multi-scale contrastive learning model. 
In the model, two GCN-based contrastive networks are trained at a patch-level (i.e., node--node) and a context-level (i.e., node--egonet) to catch the information in multiple graph scales.
Then, a statistical anomaly estimator computes the anomaly scores of target nodes on the basis of the patch- and context-level scores.


There were \cmmnt{only} a few studies that jointly take into account pattern mining and GNNs for graph anomaly detection.
For example, Zhao et al.~\cite{zhao2021synergistic} introduced a GCN-based framework, \emph{pattern mining and feature learning (PAMFUL)}, which employs pattern mining to guide the GCN in aggregating local information in order to catch global patterns in the sense of detecting the global and structural anomalies.

\paragraph{GAT-Based GAE Framework}

While the GCN-based framework exhibits good performance in anomalous node detection, its simple aggregation operation, which can cause the over-smoothing issue, limits the GCN's ability to extract useful features. To overcome this, GAT-based approaches have been developed as an alternative.

Earlier studies did not take into account the complex interactions between nodes and attributes due to the shallow learning architecture.
To solve this issue, Fan et al.~\cite{fan2020anomalydae} proposed a \emph{deep joint representation learning framework for anomaly detection through a dual AE (AnomalyDAE)}, which captures the complex interactions between the structure and attributes to detect global and structural anomalies.
Specifically, it consists of a GAT-based structure encoder and an attribute encoder, each employing an inner product decoder.

Moreover, to alleviate the over-smoothing issue in GCNs,
the \emph{graph attention-based AE (GATAE)} \cite{you2020gatae} embedded the input attributed graph by using multiple graph attention layers as an encoder in order to detect global and structural anomalies. An inner product decoder was used for reconstruction of the graph structure, and a decoder with the same architecture as its encoder was used to reconstruct node attributes.

Since most efforts have not yet handled the unseen nodes, the underlying model may be unnecessarily retrained for newly discovered nodes in graphs.
To address the issue, \emph{adversarial graph differentiation network (AEGIS)} \cite{ding2020inductive} detected the community anomalies
using an attention-based graph differentiation network (GDN).
Specifically, an encoder based on GDNs first embeds the graphs at node- and neighborhood-levels, and then a GDN-based decoder reconstructs the input graphs along with the anomaly ranking list.

\paragraph{Other GNN Frameworks}
Wang et al.~\cite{wang2021one} proposed \emph{one-class graph neural network (OCGNN)}, which aims at detecting the global anomalies.
OCGNN combines the representation capability of GNNs with a hypersphere learning objective function, which enables us to learn a hypersphere boundary and detect points that are outside the boundary as anomalous nodes.

Despite significant efforts, it remains an open challenge to learn suitable communities for effective anomaly detection.
Additionally, defining anomalies is nontrivial since the deviation patterns of anomalies are not easily disclosed.
To address these issues, Zhou et al.~\cite{zhou2021subtractive} introduced \emph{abnormality-aware GNN (AAGNN)}, which employs a subtractive aggregation method to decide the pattern deviation between a target node and its nearby nodes along with hypersphere learning for community anomaly detection. 
Specifically, 
by subtracting the feature of a target node from the aggregated feature in the $k$-hop neighborhood, abnormality of the node is determined for further embedding into the hypersphere space.


Due to the expensive labeling cost, there have been only a few methods designed in a supervised manner. 
Ding et al.~\cite{ding2021few} introduced a supervised GNN-based framework, \emph{few-shot network anomaly detection via cross-network meta-learning (Meta-GDN)}, 
which aims to detect the global anomalies on the target graph by using very limited knowledge of ground-truth anomalies from auxiliary graphs with the GCN-based encoder. Meta-GDN was built upon graph deviation networks (GDNs), which
first score each node by using a GCN-based anomaly score learner and then
use the deviation loss to enforce the model to give high anomaly scores to nodes whose features deviate from normal nodes significantly.

\subsubsection{Anomalous Edge Detection}\label{sec:AEDS}

Anomalous edges generally represent different or atypical interactions between nodes in a graph. Research on such anomalous edge detection in a static graph has been relatively limited. 

Duan et al.~\cite{duan2020aane} proposed \emph{anomaly aware network embedding (AANE)}, which was designed for plain graphs and implemented using the GCN-based GAE framework. 
This method adjusts the fitting loss and the ``anomaly-aware'' loss, which consists of both the deviation loss and the removal loss. The probability from the loss functions is the score of an edge. An edge with a lower probability is more likely to be an anomalous edge.

Song et al. \cite{song2021subgraph} proposed \emph{subgraph-based framework (SubGNN)} for fraud detection.
Sub-graphs near target edges are extracted and relabeled for entity independence.
The proposed relational graph isomorphism network (R-GIN) learns the features for precise fraud detection.

Zhang et al. \cite{zhang2022efraudcom} proposed a \emph{competitive graph neural network (CGNN)}-based fraud detection system (eFraudCom) to detect fraudulent behaviors on an e-commerce platform. The CGNN is a GCN-based GAE system. The eFraudCom system consists of a data processor and a fraud detector. Specifically, in the data processor, representative normal data were sampled, and a heterogeneous graph with the sampled normal data and the rest was generated; and in the fraud detector, neighbors in the graph were sampled and anomalous edges were detected by the CGNN.

\subsubsection{Anomalous Subgraph Detection}\label{sec:ASDS}
Anomalous subgraph detection is far more challenging than anomalous node or edge detection. For one noticeable thing, nodes and edges in an anomalous subgraph may be considered normal in themselves. Moreover, a subgraph can be very diverse in its structure and size. Presumably due to this challenge, there exist only a limited number of studies in the literature. We have found one GAT-based GAE method.

Huang et al.~\cite{huang2021hybrid} proposed a GAT-based GAE framework, called \emph{hybrid-order graph attention network (HO-GAT)}, to detect anomalous nodes and subgraphs simultaneously by using the structure and attributes of abnormal nodes and subgraphs. 
To define the most representative subgraph, the motif --- the widely studied higher-order structure characterized as a densely connected subgraph --- was leveraged. 
To model the relations between nodes and motifs, HO-GAT effectively measured the significance of four relations: a node to another node, a node to a motif, a motif to a node, and a motif to another motif.
Specifically, the graph attention layer was used in the encoder in order to capture information of the four relations mentioned above. Additionally, two individual decoders were employed to reconstruct the structure and attributes.

\subsubsection{Graph-level Anomaly Detection}

Graph-level anomaly detection (GLAD) finds graphs that are notably different from most of the graphs in a set of graphs.
There have been limited studies on addressing this task.
As mentioned above, GNN-based methods have proven its capabilities in graph-structured data, and several methods such as OCGNN \cite{wang2021one} and DOMINANT \cite{ding2019deep} have successfully applied GNN to graph anomaly detection tasks.
Recently, there have been attempts to adapt GNN-based graph classification methods to GLAD tasks.

Zhao and Akoglu \cite{zhao2021using} proposed \emph{one-class graph-level anomaly detector (OCGIN)} based on Graph Isomorphism Network (GIN) \cite{xu2018powerful}, which has drawn attention to barely studied GLAD tasks while improving the detection performance.
GIN \cite{xu2018powerful} generalizes the Weisfeiler-Lehman (WL) test \cite{weisfeiler1968reduction}, which can efficiently distinguish a broad class of graphs \cite{babai1979canonical}, and achieves powerful classification capability among GNNs.
Specifically, it optimizes the one-class Deep Support Vector Data Description (deep SVDD) objective function \cite{ruff2018deep} at the output layer of a GIN model.

To further improve the performance and overcome the hypersphere collapse where the deep one-class objective encourages all graph embeddings in the training data to concentrate within a hypersphere, Qiu et al. \cite{qiu2022raising} presented a new GNN-based approach, \emph{one-class graph transformation learning (OCGTL)}, which integrates advantages of deep one-class classification (deep OCC) \cite{ruff2018deep,zhao2021using} and self-supervised anomaly detection with learnable transformations \cite{qiu2021neural}.
Specifically, OCGTL consists of a set of GNNs to embed its input graphs into a latent space. Afterward, the various embeddings were used for training to be close to the reference embedding.

Ma et al.~\cite{ma2021deep} proposed a GCN method, \emph{global and local knowledge distillation (GLocalKD)}, which extracts graph- and node-level representations from a set of graphs by using two different stacked GCN modules to detect locally or globally anomalous subgraphs.
Specifically, it consists of two GCNs, namely, a random target network and a predictor network, that have the same GCN architecture and incur the same distillation losses.
GLocalKD learns the graph normality at a fine-grained level by using the predictor network to predict the graph- and node-level representations produced by the random target network.

\subsection{GNN-based Dynamic Graph Anomaly Detection}
Unlike a static graph, \emph{temporality} is an important factor in a dynamic graph whose structure or attributes change over time. 
Recently, various methods for detecting anomalies in graphs changing or evolving over time have been proposed based on graph communities, compression, decomposition, distance metrics, and probabilistic modeling of graph features~\cite{ranshous2015anomaly}.
There are several approaches proposed for dynamic graphs where GCN is combined with deep learning methods that are suitable for temporal processing, such as recurrent neural network (RNN), gated recurrent unit (GRU), and transformer. 
A few studies addressed detecting anomalies in edges or in nodes of a dynamic graph.
There is no study that addresses detecting anomalous subgraphs yet.

\subsubsection{Anomalous Edge Detection}\label{sec:AEDD}
The basic concept of using GCN and GRU together is that GCN extracts features from a graph and GRU captures historical information useful for anomaly detection.
Alongside the fact that GCNs do not consider the temporal factors in dynamic graphs,
\emph{anomaly detection in dynamic graph (AddGraph)} \cite{zheng2019addgraph} combined GCN and GRU, which enables us to integrate long-term and short-term patterns in a window in order to describe the normal edges by using structural, attribute, and temporal information.
Specifically, it leveraged GCN to process the previous node state with edges in the current graph by harnessing the structural and attribute features of nodes.
The nodes' states in a short window were used as the short-term information.
The GCN output and the shot-term information were then combined in a contextual attention-based GRU to extract the hidden state of nodes in order to calculate the anomaly probability of an edge.

Note that Zheng et al.'s work \cite{zheng2019addgraph} requires the entire set of nodes as the input and, thus, is not able to effectively manage newly added nodes. To address this issue, Zhu et al.~\cite{zhu2020flexible} proposed \emph{anomaly detection in dynamic network (DynAD)}, which uses evolving GCNs, consisting of multiple layers at each time, and attention-based GRU for adaptive parameter learning.

There are multi-level data structure (in addition to temporal factors) inherent in graphs. Thus, a hierarchical convolutional network model may be conducive to simultaneously capture global and local features from those graphs. 
In this regard, Wang et al.~\cite{wang2020hierarchical} employed a \emph{hierarchical GCN} with a Laplacian-based graph coarsening algorithm to leverage the multi-level structure. In addition, they captured temporal features by using GRU as well.  

To take structural changes around the target edge in a window into consideration, Cai et al.~\cite{cai2021structural} proposed \emph{structural graph neural network (StrGNN)}, which produces $h$-hop subgraphs from each temporal graph and uses GCN and GRU to extract features and temporal dependencies. 
Specifically, $h$-hop subgraphs centered on the target edge at each timeline were extracted firstly.
Then, GCNs and pooling technique were used to exploit features from the subgraphs. Subsequently, GRUs captured the temporal dependencies.

 

\subsubsection{Anomalous Node Detection}\label{sec:ANEDD}
Dynamic graphs often show complex variation patterns, which can be interpreted by the stochasticity and spatiotemporal relationships between nodes and edges. 
However, previous deterministic methods 
could not handle such a problem. To solve this, Yang et al. \cite{yang2020h} proposed \emph{hierarchical variational graph recurrent autoencoder (H-VGRAE)} by employing GCN-based GAE and RNN---specifically, by stacking multiple GCN layers and combining the GCN layers with dilated RNN (DRNN) layers in the encoder. The same methods in the encoder were used in the decoder except the Bernoulli and Gaussian MLP for reconstruction in order to detect anomalous nodes.
H-VGRAE jointly learns the anomaly features of the nodes and edges, and detects node anomalies using automatic threshold selection.

Zhang et al. \cite{zhang2022malware} proposed a novel \emph{dynamic evolving graph convolutional network (DEGCN)} model to capture evolving patterns of both local node-level and global graph-level software behaviors.
It consists of three stages. First, the multi-scale graphs are generated by sliding windows. Second, on a directed GCN (DGCN) model, in-degree node features and out-degree node features are summarized simultaneously, and on a graph encoding-based GRU (GGRU) model, the evolving patterns of graphs are learned for its time steps.
Third, the features from DGCN and GGRU are combined, and MLP layer calculates the anomality score.

Zola et al. \cite{zola2022network} proposed a graph-based approach to detect malicious connections on traffic networks.
Specifically, in the first stage, a temporal dissection operation was used to split the entire network information into time intervals and to extract Traffic Dispersion Graphs (TDGs).
In the second stage, new synthetic samples for each TDG were created to balance the number of attackers and normal nodes.
Finally, the TDGs were trained by GCNs and anomalous nodes were detected.

\section{Opportunities and Challenges}\label{sec:oppchal}
GNN-based graph anomaly detection is a quite challenging problem with lots of potentials to build sophisticated methods based upon the existing approaches presented in this survey, particularly for detecting edge anomalies and subgraph anomalies. Further, there are research opportunities and open challenges that are important and not addressed yet. This section summarizes some of them.

\paragraph{Explainable GNNs for Detecting Graph Anomalies} 
The output of anomaly detection methods is either an anomaly score or a top-$k$ ranking list. To intuitively understand the meaning of the score and list, additional explanations need to be provided. 
Since GNN-based methods are inherently less interpretable than traditional machine learning approaches, it is important to resolve the issue along with explainable models for anomaly detection. Currently, there is relatively little work on designing explainable GNN models for graph anomaly detection (e.g., Deng et al.~\cite{deng2021graph}).

\paragraph{Identification of Graph Anomalies with GNNs}
Anomalies are mostly recognized on the basis of a reconstruction loss and a distance-based stochastic function. 
Although this loss function is capable of capturing deviating patterns in training, the reconstruction loss is vulnerable to noise and the distance-based function does not work properly when anomalous and normal data are distant in the embedding space. These challenges will be no less, or rather worse, for the problem of identifying graph anomalies.
While many GNN-based approaches embed graphs into either Euclidean or hyperbolic spaces, they do not fully utilize the information available in graphs or lack the flexibility to model intrinsic complex graph geometry~\cite{zhu2020graph}. Therefore, it is a promising future research direction to design a suitable embedding space and loss functions to detect graph anomalies.

\paragraph{Class Imbalance in Graph Anomaly Detection with GNNs}
Imbalance between normal and anomalous data is inevitable since the anomalies tend to occur rarely. As the model performance is heavily dependent on training data, the class imbalance is a challenge that must be overcome. 
Rather than na\"{i}vely using negative sampling, a better strategy may be augmenting the training data, which is also a challenging issue in graph domains. There is little research carried out to resolve the issue of class imbalance in graph anomaly detection with GNNs in the literature (e.g., Zhao et al.~\cite{zhao2021data}).

\paragraph{Anomaly Detection with GNNs from Heterogeneous Graphs}
A heterogeneous graph is a graph that contains multiple types of nodes and edges and often occurs in practice. Studies on heterogeneous graph anomaly detection with GNNs have been largely under-explored in the literature. Handling the heterogeneity in solving the anomaly detection problem along with GNNs would be challenging, particularly if compounded by the temporal factors in a dynamic graph, due to the modeling difficulty including extraction of multiple relations of nodes/edges. 
Feasible strategies may include (1) sampling target edges or subgraphs via meta-paths and (2) extracting useful information via embedding the types of heterogeneous graphs.

\paragraph{Few-Shot Graph Anomaly Detection with GNNs}
Although GNN approaches have rapidly advanced over the past a few years, few-shot graph anomaly detection with GNNs has nor been studied much yet. While, in real-world scenarios, it is easy to \cmmnt{acquired}obtain a few labeled anomalies from graphs similar to the target graph, there have been only a few recent attempts to leverage such a potential to develop GNN models that can leverage a small number of labeled anomalies (e.g., Ding et al.~\cite{ding2021few}). There are still open challenges on designing diverse meta-learning algorithms to conduct few-shot graph anomaly detection aided by GNNs.

\section{Conclusion}
In this paper, we provided a comprehensive survey of state-of-the-art graph anomaly detection methods that were built upon GNN models. Additionally, we presented several opportunities and challenges for further research in this area. 
A majority of research efforts have been concentrated on detecting node anomalies from static graphs, while types of graphs and anomaly types are rapidly expanding to include dynamic graphs and edge/subgraph/graph-level anomalies. This trend is expected to continue in the coming years.
\textcolor{black}{Some specific application domains, such as anomaly detection in IoT, in program analysis (e.g., Android malware and OS verification), remain for future research.}
We hope that this survey will trigger and stimulate more active research on graph anomaly detection using GNNs and be a stepping stone for subsequent surveys as the research progresses.

\section*{Acknowledgment}
This research was supported by NRF grant funded by the Korea Government (MSIT) (No. 2021R1A2C3004345) and by IITP grant funded by the Korea Government (MSIT) (No. RS-2022-00155857, Artificial Intelligence Convergence Innovation Human Resources Development (Chungnam National University)) and BK21 FOUR Program by Chungnam National University Research Grant, 2022.

\bibliographystyle{IEEEtran}
\bibliography{mybibfile.bib}

\end{document}